\documentclass{article}

\usepackage{arxiv}

\usepackage[utf8]{inputenc} 
\usepackage[T1]{fontenc}    
\usepackage{hyperref}       
\usepackage{url}            
\usepackage{booktabs}       
\usepackage{amsfonts}       
\usepackage{nicefrac}       
\usepackage{microtype}      
\usepackage{lipsum}
\usepackage{graphicx}
\usepackage{amsmath}
\usepackage{adjustbox}
\usepackage{multirow} 
\usepackage{authblk}
\usepackage{titlesec}

\graphicspath{ {./images/} }

\title{Towards Reliable Time Series Forecasting under Future Uncertainty: Ambiguity and Novelty Rejection Mechanisms}
\author[1]{Ninghui Feng*}
\author[1]{Songning Lai*}
\author[1]{Xin Zhou}
\author[1]{Jiayu Yang}
\author[2]{Kunlong Feng}
\author[1]{Zhenxiao Yin}
\author[1]{Fobao Zhou}
\author[3]{Zhangyi Hu}
\author[1]{Yutao Yue}
\author[1]{Yuxuan Liang}
\author[4]{Boyu Wang}
\author[1]{Hang Zhao}

\affil[1]{HKUST(GZ), \texttt{\{ninghuifeng, songninglai, xinzou, jyang729, fobaoz, yutaoyue, yuxliang, hangzhao\}@hkust-gz.edu.cn}}
\affil[1]{HKUST(GZ), \texttt{zyin368@connect.hkust-gz.edu.cn}} 
\affil[2]{Qingdao University, \texttt{fengkunlong@qdu.edu.cn}}
\affil[3]{Wuhan University, \texttt{zhangyi\_hu@whu.edu.cn}} 
\affil[4]{The University of Western Ontario, \texttt{bwang@csd.uwo.ca}}
\begin{document}
\maketitle
\begin{abstract}
In real-world time series forecasting, uncertainty and lack of reliable evaluation pose significant challenges. Notably, forecasting errors often arise from underfitting in-distribution data and failing to handle out-of-distribution inputs. To enhance model reliability, we introduce a dual rejection mechanism combining ambiguity and novelty rejection. Ambiguity rejection, using prediction error variance, allows the model to abstain under low confidence, assessed through historical error variance analysis without future ground truth. Novelty rejection, employing Variational Autoencoders and Mahalanobis distance, detects deviations from training data. This dual approach improves forecasting reliability in dynamic environments by reducing errors and adapting to data changes, advancing reliability in complex scenarios.
\end{abstract}


\section{Introduction}
In recent years, time series forecasting has been widely used in finance \cite{sezer2020financial}, medicine \cite{morid2023time}, energy \cite{rajagukguk2020review}, environment \cite{hewage2020temporal} and other fields \cite{lim2021time, lai2024ftsframeworkfaithfultimesieve}. With the increase of data scale and computing power, deep learning has gradually become the mainstream of time series modeling \cite{zhou2021informer, feng2024timesieveextractingtemporaldynamics}. It can use massive historical data to capture complex time series patterns, bringing higher reliability and flexibility to practical applications.

However, time series forecasting, particularly in environmental management, faces significant challenges in predicting changes in air quality or hazard levels \cite{cheng2022intelligent}, while inaccurate forecasts can delay emergency interventions and heighten public health risks. The inherent uncertainty of future data, which lacks real ground truth for evaluation, complicates the assessment of a model's performance \cite{wen2022robust,lai2023faithful}. Traditional evaluation metrics, such as Mean Squared Error (MSE) and Mean Absolute Error (MAE) \cite{adhikari2013introductory}, rely on known labeled data, making it difficult to validate predictions against unobservable future outcomes. This discrepancy between historical performance and actual predictive accuracy is further exacerbated by the fact that models, while potentially reliable on past data, may falter under unforeseen future conditions, leading to substantial risks. In finance, erroneous forecasts can result in considerable economic losses \cite{elegbede2022evaluation}; in industrial control systems, inaccurate predictions could lead to equipment failures or safety hazards \cite{amodei2016concrete}. Thus, there is an urgent need to develop methods capable of evaluating the reliability and accuracy of time series forecasting models in the face of unknown futures. In these scenarios, We believe that increasing a model’s confidence (i.e., reducing prediction uncertainty) can significantly enhance the reliability of its forecasting results.

In classification tasks, researchers have tackled similar challenges by introducing “selective classification” or “refusal mechanisms” to lessen the severe consequences of misclassification in critical scenarios. Although some work \cite{denis2021regressionrejectoptionapplication} has extended refusal mechanisms to regression tasks (e.g., kNN-based methods for non-time-series data), these traditional models lack the ability to handle large-scale, high-dimensional, and complex dependencies, thus failing to meet the requirements of time-series forecasting \cite{taunk2019brief}. Meanwhile, other traditional methods with built-in confidence outputs (such as Gaussian Processes) generally cannot match deep learning in terms of representational capacity, so they also encounter bottlenecks in complex scenarios \cite{chauhan2018review}.

\begin{figure}[tbp]
  \centering
  \includegraphics[width=0.6\linewidth]{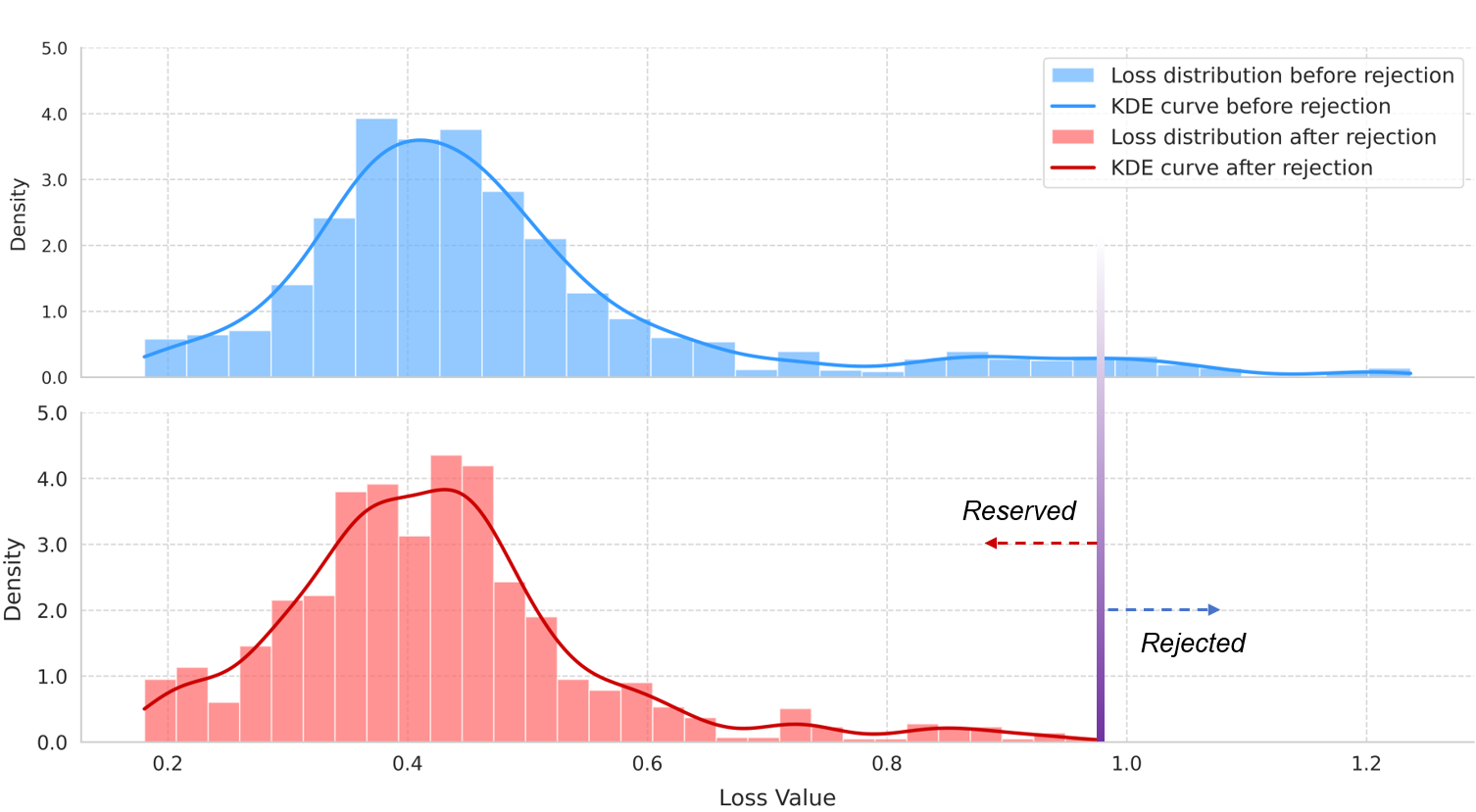}
  \caption{Distribution of sample losses predicted by the model. The vertical axis represents the density of the distribution, while the horizontal axis shows the specific values of the losses. The upper part displays the loss distribution without rejection, and the lower part shows the loss distribution including the rejection.}
 
  \label{fig:rs_2}
  \vspace{-5ex}
\end{figure}

By contrast, deep learning relies on end-to-end feature extraction and large-scale parameter optimization. However, its prediction errors primarily stem from two sources \cite{hendrickx2024machine}: first, the model may underfit on in-distribution samples, leading to biased predictions \cite{taori2023data,shao2020understanding,lakkaraju2016discovering}
; second, for out-of-distribution (OOD) samples, the model often lacks sufficient experience to predict them accurately \cite{zhang2021understanding,soen2024rejection,ramalho2020density}. Since kNN-based refusal mechanisms focus on local neighborhood variance, they do not fit neatly into the structure and training approach of deep neural networks, making them difficult to apply directly to high-dimensional and dynamic time-series forecasting \cite{denis2021regressionrejectoptionapplication,benidis2022deep,sezer2020financial,makridakis2023statistical}.

In light of these challenges, there is an urgent need for a novel approach that can effectively assess prediction uncertainty without relying on future ground truth, thereby enhancing the reliability of models in time series forecasting tasks. To this end, we introduce a dual rejection mechanism comprising ambiguity rejection and novelty rejection. Specifically, novelty rejection is first employed to detect out-of-distribution samples. If the sample is identified as in-distribution, ambiguity rejection is then applied to assess whether the model is underfitting. By estimating these distinct sources of uncertainty within a unified framework, the model can first determine whether a sample is novel and subsequently evaluate its confidence level. This allows the model to refrain from making predictions when the confidence is low, thus reducing the risk of errors.

In the training data, leading to potential inaccuracies in predictions. Novelty rejection addresses the detection of inputs that differ significantly from the training data distribution, enabling the model to identify and reject novel or ood samples. This combined approach aims to improve the model's reliability by preventing it from making predictions when confidence is lacking or when faced with unfamiliar data. The main contributions of this paper are as follows:

\noindent \textbf{(i) Integrated Rejection Mechanism for Time Series Forecasting:} We introduce an innovative rejection mechanism that combines ambiguity and novelty detection specifically tailored for time series forecasting within deep learning frameworks. This approach seamlessly integrates with end-to-end neural network training, enabling the model to handle large-scale, high-dimensional data while identifying and rejecting predictions with uncertain or out-of-distribution inputs.

\noindent \textbf{(ii) Confidence Estimation Under Uncertainty:} Addressing the challenge of evaluating forecast reliability without future labels, our method quantifies uncertainty by leveraging historical prediction error variance. This allows the model to abstain from predictions when uncertainty is high, thereby preventing potential severe errors and enhancing overall forecast reliability.

\noindent \textbf{(iii) Enhancing Reliability Through Statistical Measures:} To improve the model's robustness against concept drift and data shifts, we implement a novelty rejection system using Variational Autoencoders (VAEs) and Mahalanobis distance. This ensures that the model can reliably detect and respond to significant deviations from the training data distribution, further solidifying its predictive accuracy over time.

\begin{figure*}[tbp]
  \centering
  \includegraphics[width=0.9\textwidth]{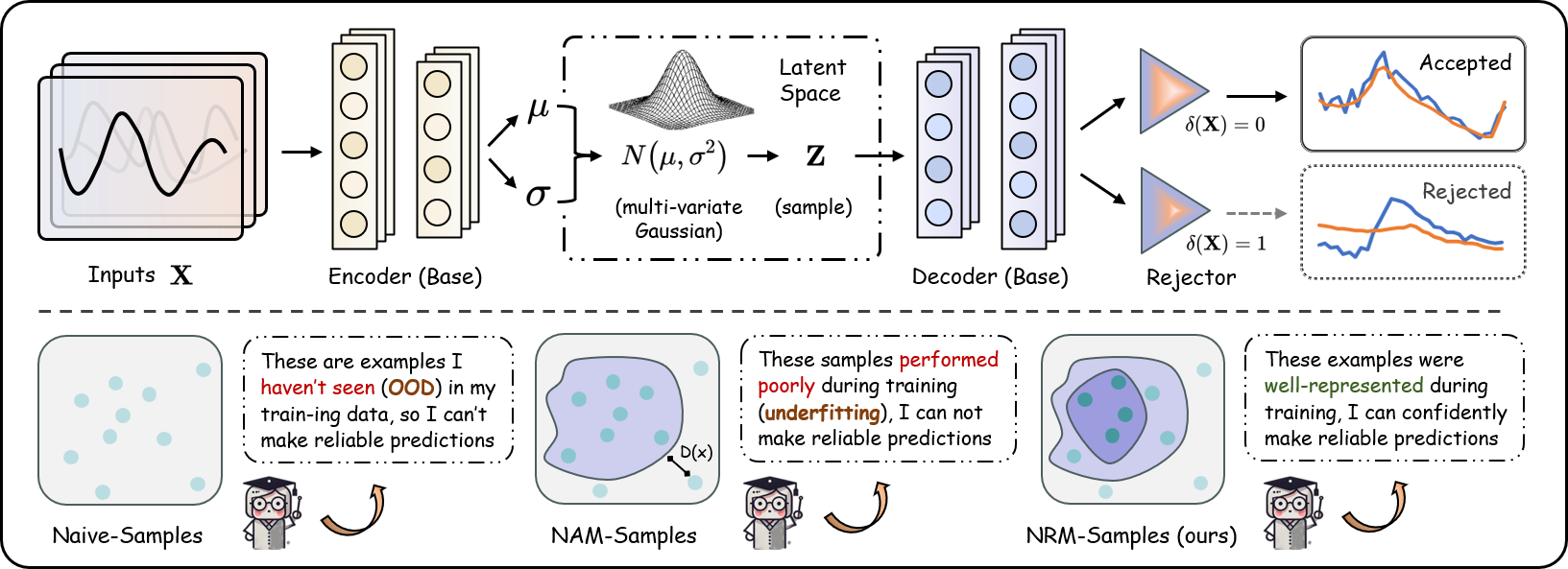} 
  \caption{Flowchart of the model. Here, $D(x)$ represents the Mahalanobis distance, $\delta(\mathbf{X})$ is the indicator function. When a sample is fed into the model, it first measures the distance between the sample and the high-dimensional distribution learned by the model. If the distance is large, the model chooses to reject the prediction for that sample. If the distance is small, the sample is passed to the rejector, which evaluates whether the model can make an accurate prediction. If the model cannot make an accurate prediction, the sample is rejected.}
  \label{fig:model}
\end{figure*}
\section{Preliminaries and Related Works}

\label{prw}
In this section, we introduce the basic definitions of multivariate time series prediction tasks and classification tasks with rejection options, laying the foundation for the subsequent methods.

\subsection{Multivariate Time Series Prediction Task}

Multivariate time series prediction aims to forecast future values based on historical observation data \cite{wei2019multivariate}. Given historical data $\mathbf{X} = \{\mathbf{x}_1, \mathbf{x}_2, \ldots, \mathbf{x}_T\} \in \mathbb{R}^{T \times N}$, where $T$ denotes the time steps and $N$ represents the number of variables, our goal is to learn a prediction model $f$ to forecast the values for the next $S$ time steps:

\begin{equation}
\mathbf{Y} = \{\mathbf{x}_{T+1}, \mathbf{x}_{T+2}, \ldots, \mathbf{x}_{T+S}\} = f(\mathbf{X}),
\end{equation}
 
\noindent here, $\mathbf{x}_t \in \mathbb{R}^N$ represents the observations of all variables at time step $t$.

It is important to note that in practical applications, different variables may have systematic time lags, leading to differences in physical measurements and statistical distributions among elements in $\mathbf{x}_t$. This increases the complexity and challenges of time series prediction.

\subsection{Classification Tasks with Rejection Options}
In traditional classification tasks, the input space is denoted as $\mathcal{X} \subseteq \mathbf{X}$, and the output space is $\mathcal{Y} = \{1, 2, \ldots, K\}$, where $K$ represents the number of classes. The objective of a classification model is to learn a mapping function \(f: \mathcal{X} \rightarrow \mathcal{Y}\) that allows the model to accurately predict the corresponding class \(y \in \mathcal{Y}\) for any given input \(\boldsymbol{x} \in \mathcal{X}\).

However, in some high-risk application scenarios, incorrect classifications can lead to severe consequences. To address this, classification strategies with rejection options have been introduced, allowing the model to refuse to make predictions when it lacks confidence in certain inputs, thereby reducing overall risk.

The rejection loss function is defined as:

\begin{equation}
\ell_{0\text{-}1\text{-}c}(f(\boldsymbol{x}), y) = \begin{cases}
c, & \text{if } f(\boldsymbol{x}) = \otimes, \\
\mathbb{I}[f(\boldsymbol{x}) \neq y], & \text{otherwise},
\end{cases}
\end{equation}
 
\noindent where $\otimes$ denotes the rejection of the prediction, $c \in (0, 0.5)$ is a fixed rejection cost, and $\mathbb{I}[\cdot]$ is the indicator function, which takes the value 1 when the condition is true and 0 if it is false.

The expected risk can be expressed as:

\begin{equation}
R(f) = \mathbb{E}_{(\boldsymbol{x}, y) \sim \mathbb{P}(\boldsymbol{x}, y)} [\ell_{0\text{-}1\text{-}c}(f(\boldsymbol{x}), y)].
\end{equation}
 
Under this setting, the optimal rejection strategy (i.e., Chow’s decision rule \cite{chow1970optimum}) is to reject the prediction when the maximum predicted probability for an input is less than $1 - c$:

\begin{equation}
    \arg \min_{f} R(f) = \arg \min_{f} \mathbb{E}_{(\boldsymbol{x}, y) \sim \mathbb{P}(\boldsymbol{x}, y)} [\ell_{0\text{-}1\text{-}c}(f(\boldsymbol{x}), y)].
\end{equation}

\section{Method}

In this section, we propose a novel framework that incorporates both ambiguity rejection and novelty rejection mechanisms into multivariate time series forecasting tasks. This approach addresses the challenges associated with the absence of future ground truth and the continuous nature of regression problems. We begin by defining the problem and then introduce each rejection mechanism in detail.

In high-risk applications, inaccurate predictions can have severe repercussions. To address this, we propose a framework that integrates rejection options into multivariate time series forecasting. Expanding upon the conventional predictor $f$, which forecasts future values directly, we introduce a reject option $\Gamma$. This allows the predictor to either produce a forecast or abstain from making one, based on predefined criteria. To balance prediction accuracy with the frequency of rejections, we establish a risk function that considers both the error in predictions and the rate of rejections.

\begin{equation} \mathcal{R}_{\lambda}(\Gamma) = \mathbb{E}\left[ \ell(f(\boldsymbol{x}), \boldsymbol{y}) \cdot \mathbb{I}(\Gamma(\boldsymbol{x}) \neq \emptyset) \right] + \lambda \cdot \mathbb{P}(\Gamma(\boldsymbol{x}) = \emptyset), \end{equation}

\noindent where \( \ell \) is the loss function (such as mean squared error), \( \mathbb{I} \) is the indicator function, and \( \lambda \geq 0 \) is a parameter that controls the trade-off between prediction error and rejection rate. Our goal is to find the optimal predictor \( \Gamma^* \) that minimizes this risk function:

\begin{equation} 
\Gamma^{*} = \arg\min_{\Gamma} \mathcal{R}_{\lambda}(\Gamma).
\end{equation}

By incorporating the reject option, our model can avoid making high-risk predictions, thereby enhancing overall reliability and performance in time series forecasting tasks.

\subsection{Rejection Mechanism in Multivariate Time Series Prediction}

In section, We first describe the ambiguity rejection mechanism, which relies on the variance of prediction errors to measure the model's confidence in its predictions. To accurately identify when to reject a prediction, we measure the confidence of the sample by constructing prediction intervals using the standard method based on the Student's t-distribution confidence interval \cite{goutis1992increasing}, and further demonstrate the positive correlation between the variance of prediction errors and confidence.

Assuming we have obtained the predicted values $\hat{y}_i$ and the true values $y_i$ for $i = 1, 2, \dots, N$, the variance of the prediction error can be expressed as:

\begin{equation} 
\text{Var}(y - \hat{y}) = \frac{1}{N-1} \sum_{i=1}^{N} (y_i - \hat{y}_i)^2 .
\end{equation}
 
This error variance is calculated on the basis of a set of prediction errors from the training data. We believe that the variance of the prediction error is inversely correlated with the confidence level of the prediction. Specifically, a larger prediction error variance indicates a higher uncertainty in the prediction, leading to a lower confidence.

To further explore this relationship, we construct the prediction interval using the t-distribution. The confidence interval for the predicted mean at a confidence level of $1 - \alpha$ is given by:

\begin{equation} \hat{y} \pm t_{\alpha/2, N-1} \cdot SE , \end{equation}
 
\noindent where $t_{\alpha/2,  N-1}$ is the critical value from the t-distribution with $N-1$ degrees of freedom at significance level $\alpha$, and $SE$ is the standard error of the prediction, calculated as:

\begin{equation} SE = \frac{s}{\sqrt{N}}, \end{equation}
 
\noindent with $s = \sqrt{\text{Var}(y - \hat{y})}$.

By rearranging the confidence interval formula, we can express the standard error in terms of the confidence interval width $W$ (where $W = 2 \cdot t_{\alpha/2, N-1} \cdot SE$):

\begin{equation} SE = \frac{W}{2 \cdot t_{\alpha/2, N-1}}. \end{equation}


Substituting back to find the variance:

\begin{equation} \text{Var}(y - \hat{y}) = \left( \frac{W}{2 \cdot t_{\alpha/2, N-1}} \right)^2 .\end{equation}

From the above analysis, we observe that as the variance of the prediction error increases, the confidence interval widens (since $W$ is directly proportional to $SE$ and hence to the square root of the variance), indicating a decrease in confidence. This provides a theoretical basis for our proposed rejection strategy.


In traditional classification, following Chow's rule \cite{chow1970optimum}, a model rejects a prediction if the highest prediction probability is below a set threshold. For time series forecasting, lacking discrete class probabilities, we use the variance of prediction errors to gauge uncertainty. A large error variance indicates high uncertainty and low model confidence, leading to the rejection of the prediction.

We propose a rejection strategy based on the variance of the prediction error. The core idea of this strategy is that when the estimated variance of the prediction error exceeds a certain threshold determined by the desired confidence level, the model should reject the prediction. The rejection condition can be expressed as:

\begin{equation} \text{Var}(y - \hat{y}) > \left( \frac{W}{2 \cdot t_{\alpha/2, N-1}} \right)^2, \end{equation}
 
\noindent here, $\alpha$ is the significance level, and $1 - \alpha$ is the confidence level. The term $t_{\alpha/2, N-1}$ is the critical value from the t-distribution with $N - 1$ degrees of freedom. By selecting an appropriate $\alpha$, we set the acceptable level of uncertainty. If the estimated variance exceeds this threshold, it implies that the prediction is too uncertain to be reliable, and thus the model rejects the prediction.

In this context, $\text{Var}(y - \hat{y})$ is the estimated variance of the prediction error, calculated from the training data. Since future ground truth values are unavailable during testing, we assume that the variance of the prediction error for new samples can be estimated based on the model's prediction uncertainty or other internal metrics.


In time series forecasting, the rejection mechanism dynamically decides to accept or reject a prediction based on the estimated variance of the prediction error. This approach mirrors the rejection strategy in classification but focuses on prediction uncertainty instead of posterior probabilities.

Based on this, the rejection strategy in time series forecasting can be formulated similarly to the rejection mechanism in classification tasks, as follows:

\begin{equation} 
\ell_{\text{seq}-c}(f(\boldsymbol{x}), y, \delta(\boldsymbol{x})) = 
\begin{cases} 
c, & \text{if } \delta(\boldsymbol{x}) = 1, \\ \ell(f(\boldsymbol{x}), y), & \text{if } \delta(\boldsymbol{x}) = 0, 
\end{cases} 
\end{equation}
 
\noindent here, $\delta(\boldsymbol{x})$ is a binary function used to decide whether the sample should be rejected. 

When $\delta(\boldsymbol{x}) = 1$, it means that the predicted uncertainty for the sample is too high, and the model rejects the prediction, incurring a fixed rejection cost $c$.
When $\delta(\boldsymbol{x}) = 0$, it means that the predicted uncertainty for the sample is acceptable, and the model proceeds to make a prediction, using the loss function $\ell(f(\boldsymbol{x}), y)$, such as mean squared error (MSE) or other suitable loss functions for time series forecasting.
In time series forecasting tasks, $\delta(\boldsymbol{x})$ can be defined based on the estimated variance of the prediction error. Specifically:

\begin{equation}
\delta(\boldsymbol{x}) = 
\begin{cases} 
1, & \text{if } \text{Var}(y - \hat{y}) > \left( \frac{W}{2 \cdot t_{\alpha/2, N-1}} \right)^2, \\ 
0, & \text{if } \text{Var}(y - \hat{y}) \leq \left( \frac{W}{2 \cdot t_{\alpha/2, N-1}} \right)^2, 
\end{cases} 
\end{equation}
 
\noindent where $\text{Var}(y - \hat{y})$ is the estimated variance of the prediction error, $W$ is the confidence interval width, calculated as $W = 2 \cdot t_{\alpha/2, N-1} \cdot SE$, where $SE$ is the standard error and $t_{\alpha/2, N-1}$ is the t-distribution critical value at significance level $\alpha/2$ with $N - 1$ degrees of freedom.

By determining the rejection threshold through the confidence interval, we establish a direct relationship between the acceptable level of prediction uncertainty and the variance of the prediction error. If the variance exceeds this threshold, it indicates that the confidence interval is too wide (i.e., uncertainty is too high), and the model should reject the prediction to maintain reliability.

This approach ensures that predictions are only made when the model's uncertainty is within acceptable bounds, enhancing the reliability of the time series forecasting model in scenarios where future ground truth data is unavailable.

So far, we have described the ambiguity rejection mechanism. Now, we introduce the novelty rejection mechanism to further enhance the model's reliability by detecting and rejecting novel inputs that differ significantly from the training data.

Rejection mechanisms rely solely on error variance to determine whether a sample should be rejected. While these methods are useful, they may fail to capture potential deviations in the input data's distribution. To overcome this limitation, we introduce Variational Autoencoders (VAE) and Mahalanobis distance as tools to evaluate the degree of deviation of a sample from the learned data distribution. This approach allows us to better assess the novelty of the sample and improve the rejection strategy. As illustrated in Figure \ref{fig:model_flow}, the model first measures the distance between a sample and the learned high-dimensional distribution. If the distance is large, the sample is rejected, while if the distance is small, the model evaluates the prediction accuracy before deciding whether to reject the sample.

To identify novelty in the input data, we use VAE to model the underlying data distribution. The VAE achieves this by maximizing the likelihood function, or equivalently, by minimizing the following loss function:

\begin{equation}
\begin{aligned}
\mathcal{L}_{\text{VAE}}(\boldsymbol{x}_i; \phi, \theta) &= \mathbb{E}_{q_{\phi}(\mathbf{z}_i|\boldsymbol{x}_i)} \left[ -\log p_{\theta}(\boldsymbol{x}_i|\mathbf{z}_i) \right] \\
&\quad + \text{KL}\left( q_{\phi}(\mathbf{z}_i|\boldsymbol{x}_i) \parallel p(\mathbf{z}_i) \right),
\end{aligned}
\end{equation}

\noindent here, $ q_{\phi}(\mathbf{z}_i|\boldsymbol{x}_i) $ is the encoder with parameters $ \phi $ approximating the true posterior distribution, $ p_{\theta}(\boldsymbol{x}_i|\mathbf{z}_i) $ is the decoder with parameters $ \theta $ generating the reconstructed samples, $ p(\mathbf{z}_i) $ is the prior distribution (typically set to a standard normal distribution $ \mathcal{N}(0, \mathbf{I}) $), and $\text{KL}(\cdot \parallel \cdot) $ denotes the Kullback-Leibler divergence measuring the difference between the two distributions.

To further quantify the deviation of input samples from the training data distribution in the latent space, we compute the Mahalanobis distance. During training, we estimate the latent distribution $ q_{\phi}(\mathbf{z}) $ of the training data, with its mean and covariance defined as:

\begin{equation}
\begin{aligned}
\boldsymbol{\mu}_{\text{train}} &= \frac{1}{N} \sum_{i=1}^{N} \boldsymbol{\mu}_{\phi}(\boldsymbol{x}_i), \\
\boldsymbol{\Sigma}_{\text{train}} &= \frac{1}{N} \sum_{i=1}^{N} \left[ \boldsymbol{\sigma}_{\phi}^2(\boldsymbol{x}_i) \right. \allowbreak + \\
&\left. \left( \boldsymbol{\mu}_{\phi}(\boldsymbol{x}_i) - \boldsymbol{\mu}_{\text{train}} \right) \left( \boldsymbol{\mu}_{\phi}(\boldsymbol{x}_i) - \boldsymbol{\mu}_{\text{train}} \right)^\top \right],
\end{aligned}
\end{equation}
 
\noindent where $ \boldsymbol{\mu}_{\phi}(\boldsymbol{x}_i) $ and $\boldsymbol{\sigma}_{\phi}^2(\boldsymbol{x}_i) $ are the mean and variance outputs of the encoder for sample $i $. These statistics characterize the distribution of the training data in the latent space.

For a test sample $ \boldsymbol{x}_i $, with latent representations $\boldsymbol{\mu}_{\phi}(\boldsymbol{x}_i) $ and $ \boldsymbol{\sigma}_{\phi}^2(\boldsymbol{x}_i) $, we calculate its Mahalanobis distance from the training distribution as follows:

\begin{equation}
D_M(\boldsymbol{x}_i) = \sqrt{ \left( \boldsymbol{\mu}_{\phi}(\boldsymbol{x}_i) - \boldsymbol{\mu}_{\text{train}} \right)^\top \boldsymbol{\Sigma}_{\text{train}}^{-1} \left( \boldsymbol{\mu}_{\phi}(\boldsymbol{x}_i) - \boldsymbol{\mu}_{\text{train}} \right) }.
\end{equation}

A large $ D_M(\boldsymbol{x}_i) $ indicates that the latent representation of $\boldsymbol{x}_i $ deviates from the training data distribution, potentially indicating novelty errors. By assessing the Mahalanobis distance, we can effectively evaluate the novelty characteristics of input samples and improve the rejection strategy.

Building upon the existing loss function framework, we incorporate the novelty rejection mechanism using the Mahalanobis distance. Specifically, we define an updated decision function $\delta_{\text{novelty}}(\boldsymbol{x})$ 
that considers both ambiguity and novelty:

\begin{equation} 
\delta_{\text{total}}(\boldsymbol{x}) = \begin{cases} 
1, & \text{if } \delta_{\text{ambiguity}}(\boldsymbol{x}) = 1 \text{ or } \delta_{\text{novelty}}(\boldsymbol{x}) = 1, \\ 0, & \text{otherwise}, 
\end{cases} 
\end{equation}
 
\noindent where $\text{if } \delta_{\text{ambiguity}}$ is the ambiguity rejection decision based on the variance of the prediction error, as previously defined, and $\delta_{\text{novelty}}(\boldsymbol{x})$ is the novelty rejection decision based on the Mahalanobis distance:
\begin{equation} \delta_{\text{novelty}}(\boldsymbol{x}) = \begin{cases} 1, & \text{if } D_M(\boldsymbol{x}) > D_{\text{threshold}}, \\ 0, & \text{if } D_M(\boldsymbol{x}) \leq D_{\text{threshold}}. \end{cases} \end{equation}
\noindent here, $D_{\text{threshold}}$is a predefined threshold that determines when a sample is considered novel based on its Mahalanobis distance.

\subsection{Performance Bounds}

To rigorously evaluate the effectiveness of our proposed rejection strategy, we establish both upper and lower bounds on the overall risk. For the purpose of discussion, we assume that \( \lambda \) represents the cost incurred when a prediction is rejected. 
In addition, we follow the approach in \cite{denis2021regressionrejectoptionapplication} by fixing a predetermined rejection rate \( \varepsilon \) in the regression setting.

\subsubsection{Upper Bound: Ideal Rejection Strategy}
 
First, we consider the performance upper bound, which corresponds to an ideal rejection strategy. In this scenario, the model accurately identifies and rejects the \( \varepsilon \) proportion of samples with the highest prediction errors. Let \( \mathcal{R}_{\lambda}(\Gamma_{\text{no-reject}}) = L_{\text{all}} \) denote the average loss across all samples when no rejection is performed. Under this ideal strategy, the overall risk is:

\begin{equation}
\mathcal{R}_{\lambda}(\Gamma_{\text{opt}}) 
= 
(1 - \varepsilon)\cdot L_{\text{accepted}} 
\;+\;
\lambda \cdot \varepsilon,
\end{equation}
 
\noindent where \( L_{\text{accepted}} \) is the average loss of the accepted samples, satisfying \( L_{\text{accepted}} \leq L_{\text{all}} \). Because the rejected samples are precisely those with the highest errors, \( L_{\text{accepted}} \) achieves the lowest possible value, thus minimizing the overall risk.

\subsubsection{Lower Bound: Random Rejection Strategy}

Conversely, we examine the performance lower bound established by a random rejection strategy. In this case, the model rejects samples randomly, without regard to their prediction errors. The overall risk under random rejection is defined as:

\begin{equation}
\mathcal{R}_{\lambda}(\Gamma_{\text{random}}) 
= 
(1 - \varepsilon)\cdot L_{\text{all}} 
\;+\;
\lambda \cdot \varepsilon.
\end{equation}
 
\noindent here, since rejection is random, the average loss of the accepted samples remains \( L_{\text{all}} \), as there is no selective removal of high-error points.

\subsubsection{Comparison and Worst-Case Scenario}
 
By comparing the two strategies, we observe that:

\begin{equation}
\begin{aligned}
\mathcal{R}_{\lambda}(\Gamma_{\text{opt}}) &= (1 - \varepsilon) \cdot L_{\text{accepted}} + \lambda \cdot \varepsilon \\
&\le (1 - \varepsilon) \cdot L_{\text{all}} + \lambda \cdot \varepsilon = \mathcal{R}_{\lambda}(\Gamma_{\text{random}}),
\end{aligned}
\end{equation}

\noindent since \( L_{\text{accepted}} \leq L_{\text{all}} \). This inequality shows that the ideal strategy \(\Gamma_{\text{opt}}\) achieves a lower or equal overall risk compared to the random strategy \(\Gamma_{\text{random}}\).

To further analyze the reliability of our method, consider the worst-case scenario where the model fails to learn any meaningful patterns and ends up rejecting samples randomly. In this situation, let \( \Gamma_{\text{proposed}} \) denote our proposed strategy. Under the assumption that \(\Gamma_{\text{proposed}}\) is at least as selective as random rejection, the overall risk satisfies:

\begin{equation}
\mathcal{R}_{\lambda}(\Gamma_{\text{proposed}})
= 
(1 - \varepsilon)\cdot L_{\text{accepted}}
\;+\;
\lambda \cdot \varepsilon
\;\;\;\le\;\;\;
\mathcal{R}_{\lambda}(\Gamma_{\text{random}}).
\end{equation}
 
This holds because \( L_{\text{accepted}} \leq L_{\text{all}} \), ensuring that our proposed risk does not exceed that of random rejection. Moreover, in most practical cases, since our strategy is designed to reject higher-error samples more effectively, \( L_{\text{accepted}} \) is likely to be much lower than \( L_{\text{all}} \), yielding a further reduction of the overall risk.

By setting the rejection rate \( \varepsilon \) and adopting an intelligent refusal strategy based on prediction uncertainty and data novelty, our method guarantees that the overall risk is bounded above by the ideal rejection strategy and is no worse than the random one. Specifically, under the assumption that \( \lambda \) represents the rejection cost, we have:

\begin{equation}
    \mathcal{R}_{\lambda}(\Gamma_{\text{proposed}})
\;\le\;
\mathcal{R}_{\lambda}(\Gamma_{\text{random}}).
\end{equation}
 
In the worst-case scenario where the model does not learn any useful patterns and defaults to random rejection, our approach still maintains a risk at least as low as that of random rejection. This theoretical basis underscores the reliability and performance advantages of our proposed rejection strategy in multivariate time series forecasting.

\section{Experiments}
In this section, we evaluate the proposed dual rejection mechanism across three datasets: ETT, Weather, and Exchange Rate. Our evaluation is divided into three parts: Firstly, we present the datasets and baseline models for comparison. Secondly, we carry out comparative experiments to evaluate our method's performance relative to the baselines. Lastly, we conduct ablation studies to examine the impact of the novelty and ambiguity rejection components.

\subsection{Datasets and Baselines}

We evaluate the proposed dual rejection mechanism on three benchmark datasets: ETT \cite{zhou2021informer}, Weather \cite{zeng2023transformers}, and Exchange Rate datasets \cite{lai2018modeling}. To assess the effectiveness of our approach, we compare three baseline models: TimeXer \cite{wang2024timexerempoweringtransformerstime}, PAttn \cite{tan2024languagemodelsactuallyuseful} and Autoformer \cite{wu2021autoformer}. The experiments are conducted with prediction window lengths of 96, 192, 336, and 720 time steps to examine performance over different forecasting horizons \cite{wu2022timesnet}.

\subsection{Comparative Experiment}

In this experiment, we assess the effectiveness of the proposed dual rejection mechanism by comparing the baseline $TimeXer$, $PAttn$, and $Autoformer$ models with their enhanced versions (i.e., $TimeXer_{rej}$) across datasets: ETTh1, ETTh2, ETTm2, Weather, and Exchange. The evaluation metrics used are Mean Absolute Error (MAE) and Mean Squared Error (MSE) for different prediction window lengths (96, 192, 336, and 720) . The results are summarized in Table \ref{tab:baseline}. At the same time, in order to show the effect of our framework in the table, we will control the rejection rate to 10\%\cite{denis2021regressionrejectoptionapplication} . 

\begin{table*}[htp]
  \centering
  \caption{The table shows MAE and MSE loss values for models $TimeXer$, $PAttn$, and $Autoformer$ with and without rejection mechanisms across datasets ETTm2, ETTh1, ETTh2, Weather, and Exchange. These results highlight the effect of integrating rejection strategies on model performance.}
  \adjustbox{max size={1\textwidth}{\textheight}}{
    \begin{tabular}{c|ccccccccccccccc}
    \toprule 
    \multicolumn{2}{c}{Models} & \multicolumn{2}{c}{$TimeXer$} & \multicolumn{2}{c}{$TimeXer_{rej}$} & \multicolumn{2}{c}{$PAttn$} & \multicolumn{2}{c}{$PAttn_{rej}$} & \multicolumn{2}{c}{$Autoformer$} & \multicolumn{2}{c}{$Autoformer_{rej}$}   \\
    \midrule 
    \multicolumn{2}{c}{Metric} & MAE   & MSE   & MAE   & MSE   & MAE   & MSE   & MAE   & MSE   & MAE   & MSE   & MAE   & MSE    \\
    \midrule
    \multirow{4}[2]{*}{\rotatebox{90}{ETTm2}} 
          & 96    & 0.1686 & 0.1734 & 0.1438 & 0.1502 & 0.1756 & 0.1786 & 0.1487  & 0.1544 & 0.2153 & 0.2203 & 0.1778 & 0.1795 \\
          & 192   & 0.2341 & 0.2390 & 0.2005 & 0.2060 & 0.2415 & 0.2447 & 0.2040  & 0.2107 & 0.2712 & 0.2753 & 0.2270 & 0.2303 \\
          & 336   & 0.2939 & 0.2989 & 0.2623 & 0.2796 & 0.3033 & 0.3071 & 0.2660  & 0.2866 & 0.3233 & 0.3315 & 0.2835 & 0.2901 \\
          & 720   & 0.3952 & 0.3986 & 0.3473 & 0.3492 & 0.4033 & 0.4058 & 0.3520  & 0.3559 & 0.4194 & 0.4315 & 0.3626 & 0.3745 \\
    \midrule
    \multirow{4}[2]{*}{\rotatebox{90}{ETTh1}} 
          & 96    & 0.3942 & 0.3996 & 0.3897 & 0.3969 & 0.3932 & 0.3969 & 0.3859 & 0.3924 & 0.4393 & 0.4364 & 0.4251 & 0.4283 \\
          & 192   & 0.4435 & 0.4487 & 0.4252 & 0.4328 & 0.4484 & 0.4519 & 0.4264 & 0.4335 & 0.5389 & 0.5602 & 0.5206 & 0.5402 \\
          & 336   & 0.4837 & 0.4887 & 0.4507 & 0.4564 & 0.4857 & 0.4893 & 0.4528 & 0.4580 & 0.6051 & 0.5606 & 0.5825 & 0.5332 \\
          & 720   & 0.4797 & 0.4917 & 0.4622 & 0.4746 & 0.4801 & 0.4936 & 0.4609 & 0.4768 & 0.5646 & 0.5876 & 0.5507 & 0.5718 \\
    \midrule
    \multirow{4}[2]{*}{\rotatebox{90}{ETTh2}} 
          & 96    & 0.2888 & 0.2939 & 0.2806 & 0.2859 & 0.2937 & 0.2978 & 0.2805  & 0.2841 & 0.3361 & 0.3504 & 0.3214 & 0.3387 \\
          & 192   & 0.3680 & 0.3788 & 0.3550 & 0.3645 & 0.3714 & 0.3787 & 0.3575  & 0.3626 & 0.4261 & 0.4305 & 0.4053 & 0.4047 \\
          & 336   & 0.4168 & 0.4265 & 0.4089 & 0.4165 & 0.4180 & 0.4252 & 0.4055  & 0.4132 & 0.4482 & 0.4650 & 0.4332 & 0.4483 \\
          & 720   & 0.4297 & 0.4491 & 0.4177 & 0.4288 & 0.4203 & 0.4257 & 0.4088  & 0.4139 & 0.4440 & 0.4736 & 0.4286 & 0.4526 \\
    \midrule
    \multirow{4}[2]{*}{\rotatebox{90}{Weather}} 
          & 96    & 0.1654 & 0.1689 & 0.1593 & 0.1631 & 0.1873 & 0.1873 & 0.1738  & 0.1740 & 0.2580 & 0.2745 & 0.2467 & 0.2647 \\
          & 192   & 0.2129 & 0.2156 & 0.2040 & 0.2071 & 0.2314 & 0.2320 & 0.2179  & 0.2180 &0.2976 & 0.3253 & 0.2834 & 0.3131 \\
          & 336   & 0.2673 & 0.2703 & 0.2542 & 0.2572 & 0.2850 & 0.2852 & 0.2666  & 0.2666 & 0.3407 & 0.3696 & 0.3239 & 0.3509 \\
          & 720   & 0.3448 & 0.3467 & 0.3272 & 0.3268 & 0.3609 & 0.3609 & 0.3362  & 0.3363 & 0.4158 & 0.3962 & 0.3912& 0.3727 \\
    \midrule
    \multirow{4}[2]{*}{\rotatebox{90}{Exchange}} 
          & 96    & 0.0861 & 0.0869 & 0.0846 & 0.0838 & 0.0840 & 0.0846 & 0.0819  & 0.0843 & 0.1495 & 0.1539 & 0.1450 & 0.1529 \\
          & 192   & 0.1827 & 0.1822 & 0.1813 & 0.1804 & 0.1782 & 0.1774 & 0.1770  & 0.1763 & 0.2653 & 0.2736 & 0.2605 & 0.2702 \\
          & 336   & 0.3509 & 0.3435 & 0.3430 & 0.3410 & 0.3411 & 0.3333 & 0.3336  & 0.3254 & 0.4419 & 0.4351 & 0.4354 & 0.4279 \\
          & 720   & 0.8686 & 0.8750 & 0.8346 & 0.8473 & 0.8616 & 0.8451 & 0.8480  & 0.8254 & 1.0884 & 1.1088 & 1.0668 & 1.0966 \\
    \bottomrule
    \end{tabular}%
  \label{tab:baseline}%
  }
\end{table*}

The results in Table\ref{tab:baseline} indicate that the enhanced versions model generally achieves lower MAE and MSE values compared to the baseline across most prediction window lengths and datasets. This suggests that the rejection mechanism effectively reduces prediction errors by abstaining from uncertain forecasts.

\paragraph{Overall Performance} Across all datasets and window lengths, the $TimeXer_{REJ}$ model demonstrates consistent improvements in MAE and MSE. For example, in the ETTh1 dataset, MAE decreases from 0.4435  to 0.4252 at a window length of 192, and similar reductions are observed at other window lengths. Similarly, in the ETTm2 dataset, significant improvements are seen, with MAE reducing from 0.3952 to 0.3473 at a window length of 720.

\paragraph{Impact of Prediction Window Length} The effectiveness of the rejection mechanism appears to be more pronounced at shorter to medium window lengths (96 and 192). This could be due to the model's higher confidence and lower uncertainty in shorter-term predictions, allowing the rejection mechanism to more effectively identify and abstain from uncertain forecasts. At longer window lengths (336 and 720), while improvements are still observed, the gains are relatively smaller, potentially due to increased complexity and uncertainty in long-term forecasting.

\paragraph{Dataset Variability} Different datasets exhibit varying degrees of improvement with the rejection mechanism. The ETTh2 dataset shows substantial performance gains, indicating that the rejection mechanism is particularly effective for datasets with certain characteristics, such as lower baseline error rates. In contrast, the ETTm2 dataset shows more modest improvements, which may be attributed to its specific data properties or the presence of missing values in the table for certain window lengths.

\subsection{Ablation Experiment}

To understand the individual contributions of the novelty and ambiguity rejection modules in our proposed dual rejection mechanism, we conducted an ablation study. This analysis involves systematically removing or altering components of the model to assess their impact on overall performance. Specifically, we evaluated four configurations:

\noindent \textbf{(i) Baseline Model(Base):} The original model without any rejection mechanism.

\noindent \textbf{(ii) Novelty Rejection Only(NRO):} Incorporating only the novelty rejection module into the model.

\noindent \textbf{(iii) Ambiguity Rejection Only(ARO):} Incorporating only the ambiguity rejection module into the model.

\noindent \textbf{(iv) Dual Rejection Mechanism(DRM):} Incorporating both novelty and ambiguity rejection modules into the model.

\begin{table*}[htp]
  \centering
  \caption{Comparison of different models with and without rejection options across multiple datasets. The table presents the MAE and MSE for various prediction window lengths (96, 192, 336, and 720) on the ETTm2 and ETTh2 datasets. Models include the baseline versions of $TimeXer$, $PAttn$, and their enhanced versions with rejection mechanisms: $TimeXer_{NRO}$ (Non-Rejection Option), $TimeXer_{ARO}$ (Ambiguity Rejection Option), and $TimeXer_{DRM}$ (Dual Rejection Mechanism), as well as their counterparts for $PAttn$.}
  \adjustbox{max size={1\textwidth}{\textheight}}{
    \begin{tabular}{c|ccccccccccccccccc}
    \toprule
    \multicolumn{2}{c}{Models} & \multicolumn{2}{c}{$TimeXer_{Base}$} & \multicolumn{2}{c}{$TimeXer_{NRO}$} & \multicolumn{2}{c}{$TimeXer_{ARO}$} & \multicolumn{2}{c}{$TimeXer_{DRM}$} & \multicolumn{2}{c}{$PAttn_{Base}$} & \multicolumn{2}{c}{$ PAttn_{NRO}$} & \multicolumn{2}{c}{$ PAttn_{ARO}$}   & \multicolumn{2}{c}{$ PAttn_{DRM}$} \\
    \midrule
    \multicolumn{2}{c}{Metric} & MAE   & MSE   & MAE   & MSE   & MAE   & MSE   & MAE   & MSE   & MAE   & MSE   & MAE   & MSE   & MAE   & MSE & MAE   & MSE \\
    \midrule
    \multirow{4}[2]{*}{\rotatebox{90}{ETTm2}} 
          & 96    & 0.1686 & 0.1734 & 0.1592 & 0.1646 & 0.1540 & 0.1600 & 0.1438 & 0.1502 & 0.1756 & 0.1786 & 0.1652 & 0.1694 & 0.1584 & 0.1623 & 0.1487 & 0.1544\\
          & 192   & 0.2341 & 0.2390 & 0.2234 & 0.2284 & 0.2080 & 0.2139 & 0.2005 & 0.2060 & 0.2415 & 0.2447 & 0.2306 & 0.2347 & 0.2116 & 0.2173 & 0.2040 & 0.2107\\
          & 336   & 0.2939 & 0.2989 & 0.2787 & 0.2848 & 0.2754 & 0.2918 & 0.2623 & 0.2796 & 0.3033 & 0.3071 & 0.2883 & 0.2925 & 0.2788 & 0.2990 & 0.2660 & 0.2866\\
          & 720   & 0.3952 & 0.3986 & 0.3785 & 0.3848 & 0.3630 & 0.3634 & 0.3492 & 0.3492 & 0.4033 & 0.4058 & 0.3877 & 0.3908 & 0.3652 &0.3684 & 0.3520 & 0.3559\\
    \bottomrule
    \end{tabular}%
  \label{tab:ablation}%
  }
\end{table*}

The performance of these configurations across multiple datasets is summarized in Table \ref{tab:ablation}, where we compare the Mean Absolute Error (MAE) and Mean Squared Error (MSE) for various prediction window lengths (96, 192, 336, and 720) on the ETTm2 and ETTh2 datasets. This allows us to evaluate the impact of each rejection mechanism on model performance.

\paragraph{Impact of Novelty Rejection} Incorporating the novelty rejection module alone demonstrates a noticeable improvement in forecasting accuracy compared to the baseline model. By identifying and rejecting novel or anomalous inputs, the model reduces the likelihood of making inaccurate predictions on unfamiliar data patterns. This results in lower MAE and MSE values, indicating enhanced reliability in the model's forecasts.

\paragraph{Impact of Ambiguity Rejection} Similarly, integrating the ambiguity rejection module independently also leads to performance gains. This module assesses the uncertainty associated with each prediction and opts to reject those with high uncertainty. By doing so, the model minimizes errors that arise from ambiguous or noisy inputs, thereby improving overall prediction accuracy as reflected in reduced MAE and MSE metrics.

\paragraph{Combined Effect of Dual Rejection Mechanism} The most significant improvements are observed when both rejection modules are employed together. The dual rejection mechanism leverages the strengths of both novelty and ambiguity rejection, providing a comprehensive approach to managing different types of uncertainties in the data. This combination results in the lowest MAE and MSE values among all configurations, underscoring the effectiveness of our proposed method in enhancing the model's forecasting performance.

\subsection{Impact of Rejection Rate on Model Performance}

To further explore the relationship between the rejection rate and the accuracy of the remaining samples, we present the relationship between the rejection rate and the loss on the remaining data in Table \ref{tab:hp1_hp2_loss}. As shown in the table, when we initially introduce the rejection mechanism, the loss on the remaining samples decreases significantly, indicating that high-error samples are effectively filtered out. However, as the rejection rate continues to increase, the reduction in loss gradually diminishes and even tends to plateau. This may be because, after rejecting some samples, the errors on the remaining samples are mainly due to the model's inherent errors, and further increasing the rejection rate does not significantly improve the model's performance.To prevent an excessive number of rejected samples when using a fixed threshold of 0.5, we employed a validation set to tune hyperparameters and adjust the rejector's scoring mechanism, thereby controlling the rejection intensity.

\begin{table*}[htp]
  \centering
  \caption{The relationship between rejection rate and loss on the remaining data for the $TimeXer$ model on the $ETTm2$ dataset with varying prediction windows is shown. The first row indicates original loss values prior to rejection mechanism application. Following rows present loss values at different rejection thresholds, including corresponding MAE and rejection rates. Here, \textbf{$Var$} signifies the variance threshold, and \textbf{$D_M(\mathbf{x})$} denotes the Mahalanobis distance threshold for sample rejection.}
  \adjustbox{max size={0.95\textwidth}{\textheight}}{
    \begin{tabular}{c|cccccccccccccccc}
    \toprule 
    \multicolumn{1}{c}{Window size} & \multicolumn{4}{c}{$96$} & \multicolumn{4}{c}{$192$} & \multicolumn{4}{c}{$336$} & \multicolumn{4}{c}{$720$}  \\
    \midrule 
    \multicolumn{1}{c}{Para} & \textbf{$Var$}   & \textbf{$D_M(\mathbf{x}) $}   & \textbf{$MAE$}   & \textbf{$MAE_{Rate}$}   & \textbf{$Var$}   & \textbf{$D_M(\mathbf{x}) $}   & \textbf{$MAE$}   & \textbf{$MAE_{Rate}$} & \textbf{$Var$}   & \textbf{$D_M(\mathbf{x}) $}   & \textbf{$MAE$}   & \textbf{$MAE_{Rate}$} & \textbf{$Var$}   & \textbf{$D_M(\mathbf{x}) $}   & \textbf{$MAE$}   & \textbf{$MAE_{Rate}$}      \\
    \midrule
    \multirow{6}[2]{*}{\rotatebox{90}{ETTm2}} 
          & - & - & 0.1686 & 0\% & - & - & 0.2341 & 0\% & - & - & 0.2939 & 0\% & - & - & 0.3952 & 0\% \\
          & 0.8274 & 26.2126 & 0.1590 & 2\% & 0.9528 & 25.0697 & 0.2239 & 2\% & 0.9892 & 25.3982 & 0.2871 & 2\% & 0.9788 & 25.7091 & 0.3816 & 2\%\\
          & 0.7773 & 24.4443 & 0.1507 & 6\% & 0.9409 & 23.3349 & 0.2095 & 6\% & 0.9887 & 23.7511 & 0.2723 & 6\% & 0.9877 & 24.0036 & 0.3619 & 6\%  \\
          & 0.7165 & 23.4622 & 0.1438 & 10\% & 0.9261 & 22.4036 & 0.2005 & 10\% & 0.9883 & 22.8981 & 0.2623 & 10\% & 0.9846 & 23.0931 & 0.3473 & 10\% \\
          & 0.6393 & 22.7675 & 0.1406 & 12\% & 0.8807 & 21.7006 & 0.1929 & 12\% & 0.9877 & 22.2428 & 0.2560 & 12\% & 0.9895 & 22.4262 & 0.3380 & 12\% \\
          & 0.5492 & 22.2809 & 0.1374 & 16\% & 0.8406 & 21.1600 & 0.1890 & 16\% & 0.9869 & 21.7290 & 0.2510 & 16\% & 0.9827 & 21.9104 & 0.3338 & 16\% \\
    \bottomrule
    \end{tabular}%
  \label{tab:hp1_hp2_loss}%
  }
\end{table*}

This indicates that continuously increasing the rejection rate has a limited effect on improving the model's performance. Therefore, in practical applications, we should choose an appropriate rejection rate to balance improving accuracy and retaining data volume. A rejection rate that is too low may not sufficiently filter out unreliable predictions, while a rejection rate that is too high may lead to underutilization of data.

\section{Conclusions}

We proposed a novel framework for multivariate time series forecasting that incorporates both ambiguity rejection and novelty rejection mechanisms. The ambiguity rejection mechanism estimates prediction uncertainty based on the variance of prediction errors from the training data, allowing the model to reject predictions when uncertainty is high. The novelty rejection mechanism utilizes Variational Autoencoders and Mahalanobis distance to detect and reject inputs that significantly deviate from the training data distribution.

In practical applications, it is crucial to balance the rejection rate to enhance model reliability while retaining sufficient data. Our framework provides an effective approach to improve the reliability and reliability of time series forecasting models without relying on future ground truth data.

\bibliographystyle{unsrt}  
\bibliography{references}  






\end{document}